\documentclass[runningheads]{llncs}

\usepackage{eccv}

\usepackage{eccvabbrv}

\usepackage{graphicx}
\usepackage{booktabs}

\usepackage{multirow}
\usepackage{colortbl}
\usepackage{arydshln}
\usepackage{stfloats}
\usepackage{algorithmicx}
\usepackage{algorithm}
\usepackage{listings}
\usepackage{xcolor}
\usepackage{fix-cm}
\usepackage{algpseudocode}
\usepackage{changepage}
\usepackage{float}
\usepackage{caption}
\usepackage{placeins}

\newcommand\algcomment[1]{\def\@algcomment{\footnotesize#1}}

\usepackage[accsupp]{axessibility}  % Improves PDF readability for those with disabilities.

\usepackage{hyperref}
\hypersetup{colorlinks=true, urlcolor=purple}

\usepackage{orcidlink}

\begin{document}

\title{Silhouette-based Gait Foundation Model} 


\author{Dingqiang Ye\inst{1} \and
Chao Fan\inst{2,}\thanks{Corresponding Author.} \and
Kartik Narayan\inst{1} \and \\
Bingzhe Wu\inst{2} \and
Chengwen Luo\inst{2} \and
Jianqiang Li\inst{2} \and
Vishal M. Patel\inst{1}
}

\authorrunning{D.~Ye et al.}

\institute{Johns Hopkins University, USA \and
Shenzhen University, China 
\\
\email{dye6@jh.edu},
\email{chaofan996@szu.edu.cn},
\email{knaraya4@jh.edu}, \\
\email{wubingzheagent@gmail.com}, 
\email{\{chengwen, lijq\}@szu.edu.cn}, 
\email{vpatel36@jhu.edu}
}

\maketitle

\begin{figure}[t]
\centering
\includegraphics[width=\linewidth]{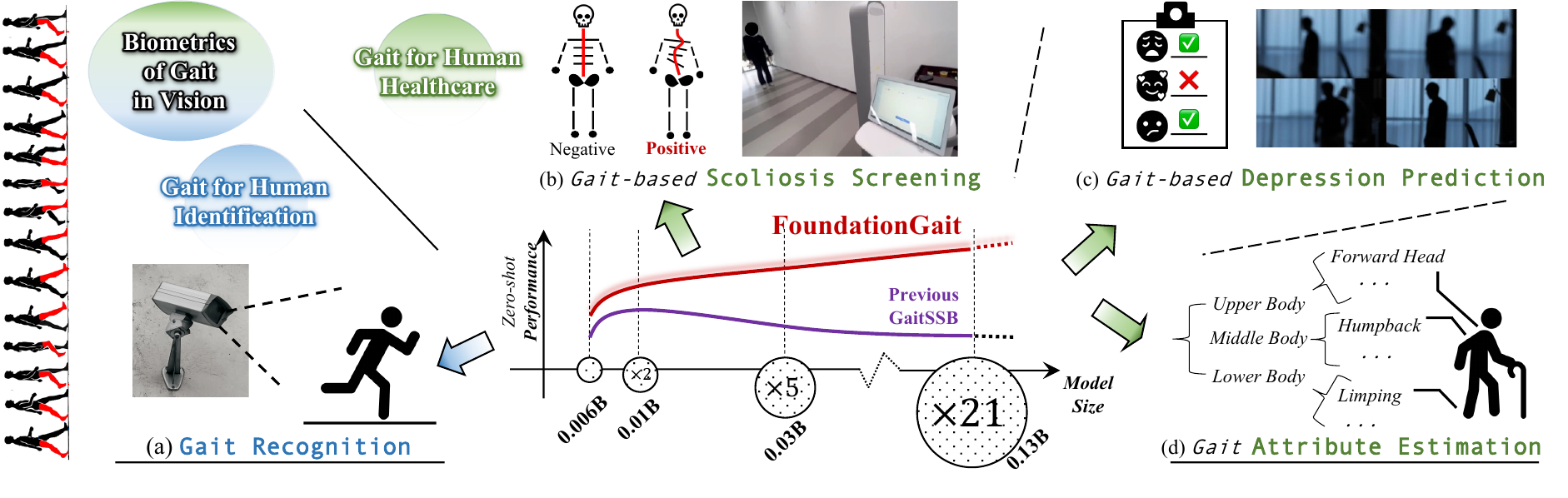}
\caption{
\textbf{FoundationGait}: A scalable and unified gait foundation model for diverse gait-related tasks, including person identification, scoliosis screening, gait attribute estimation, and depression detection.
}
\label{fig:intro}
\end{figure}

\begin{abstract}
    Gait patterns play a critical role in human identification and healthcare analytics, yet current progress remains constrained by small, narrowly designed models that fail to scale or generalize. Building a unified gait foundation model requires addressing two longstanding barriers:
    (a) Scalability – Why have gait models historically failed to follow empirical scaling trends?
    (b) Generalization – Can one model serve the diverse gait tasks that have traditionally been studied in isolation?
    We introduce \textbf{FoundationGait}, the first scalable, self-supervised pretraining framework for vision-based gait understanding.
    Its largest version has nearly 0.13 billion parameters and is pretrained on 12 public gait datasets comprising over 2 million walking sequences.
    Extensive experiments demonstrate that FoundationGait, with or without fine-tuning, performs robustly across a wide spectrum of gait datasets, conditions, tasks (e.g., human identification, scoliosis screening, depression prediction, and attribute estimation), and even input modality.
    Notably, it achieves 48.0\% \textbf{self-supervised} rank-1 accuracy on the challenging in-the-wild Gait3D dataset (1,000 test subjects) and 64.5\% on the largest in-the-lab OU-MVLP dataset (5,000+ test subjects), setting a new milestone in robust gait recognition.
    These results establish FoundationGait as a strong and versatile foundation for future gait research. 
    % All source code and models will be publicly released.
    All code and models: \url{https://github.com/ShiqiYu/OpenGait}.
    \keywords{Gait Recognition \and Gait Healthcare \and Vision Foundation Model \and Human Biometrics}
\end{abstract}

\section{Introduction}
\label{sec:intro}
Vision-based gait offer a compelling solution for large-scale, low-cost, and non-invasive applications—including human identification and healthcare~\cite{niyogi1994analyzing,nixon2010human,zhou2024gait,liu2024depression,wang2025ra,turaga2008machine}—as illustrated in Fig.~\ref{fig:intro}. Their ability to capture motion remotely and unobtrusively in unconstrained environments makes gait uniquely suitable for real-world deployment.
Yet, despite this promise, a dedicated vision-based Gait Foundation Model has not emerged, even as large vision models (LVMs)~\cite{oquab2023dinov2, kirillov2023segment, peebles2023dit} have reshaped many related domains.

Recent progress in domain-specific LVMs~\cite{guo2024skysense, gu2024AnomalyGPT, stevens2024bioclip, khirodkar2024sapiens, kim2025sapiensid}, which balance specialization with broad generalization, has demonstrated their practical value and attracted significant attention from both academia and industry~\cite{landingai2024domain}. Motivated by these developments, this paper takes a  step toward establishing such a foundation model for gait.

Recent efforts toward building a gait foundation model generally follow two directions.
The first, exemplified by BigGait~\cite{ye2024biggait} and subsequent studies~\cite{jin2025denoising, ye2025biggergait}, focuses on transferring capacity from existing LVMs—such as DINOv2~\cite{oquab2023dinov2} and Stable Diffusion~\cite{rombach2022high}—into the gait domain.
The second, represented by GaitSSB~\cite{fan2023learning} and related work~\cite{meng2025mirrorgait, wang2025ra}, instead trains gait models from scratch using vision-only or vision–language contrastive pretraining strategies~\cite{radford2021learning, chen2021exploring}.

Both directions demonstrate that rich and discriminative gait representations can indeed be learned from walking videos under supervised or self-supervised regimes. 
Despite the significant progress, they---similar to much of the longstanding gait literature~\cite{fan2025opengait, zhou2024gait, liu2024depression, wang2025ra}---remain task-specific, typically optimizing models for individual datasets rather than pursuing unified, scalable solutions.

Two core properties required for a true gait foundation model remain insufficiently explored.
(a) \textbf{Size Scalability.} Although empirical scaling trends are a known driver of progress in modern foundation models, systematic investigations of scalability for video-based gait modeling are rare. In practice, we find that many gait-specific architectures fail to benefit from increasing parameters beyond relatively small sizes ($\approx$10M); performance often saturates or even drops sharply as model capacity grows (Sec.~\ref{sec:unscale}). 
Historically, this limitation is mainly attributed to the simplicity of gait inputs—low resolution ($64 \times 44$) sparse binary silhouettes—together with limited training data, leading to overfitting and hindering large-scale models.
However, we observe that large-scale models suffering from fine-grained detail loss is also a key factor to the unscalable issue.
(b) \textbf{Cross-task Generalization.} A defining feature of foundation models is their ability to transfer across diverse downstream tasks. Yet, gait research has traditionally treated recognition-oriented tasks and healthcare-oriented analyses as separate problem spaces, despite both relying on subtle, fine-grained motion dynamics.

Motivated by these gaps, we investigate the design principles of a true gait foundation model and introduce FoundationGait, a scalable self-supervised pretraining framework that generalizes effectively across a broad spectrum of gait datasets and downstream tasks. 
Thanks to its frequent usage (62.8\% in recent top venues according to OpenGait~\cite{fan2025opengait}), FoundationGait takes the popular silhouette sequence as input. 
Our exploration centers on two key questions:

\begin{itemize}
    \item \textit{How can we scale gait models from 0.01B parameters
    to the 1B  range}?
    In Fig.~\ref{fig:unscalable}, even with a million-level diverse training data, current gait models still fail to scale.
    We find that preserving fine-grained local body-part cues is essential for scaling. Building on this insight, we develop a novel \textit{part-aware pretraining} strategy that enables FoundationGait to scale by \textbf{20$\times$} over widely used baselines (e.g., DeepGaitV2’s $\approx$0.006B backbone~\cite{fan2025opengait}), while achieving consistent performance gains across datasets and tasks.
    \item \textit{How can we build an efficiently large gait corpus for pretraining?} 
    Although the existing GaitLU-1M~\cite{fan2023learning} dataset contains nearly one million silhouette sequences, it has been criticized for its limited photographic diversity~\cite{meng2025mirrorgait}.
    To address this limitation, we curate 12 publicly available gait datasets spanning both recognition tasks~\cite{yu2006framework, takemura2018multi, song2022casia, zou2024cross, li2023depth, shen2023lidargait, zheng2022gait, zhu2021gait, fan2023learning} and healthcare applications~\cite{zhou2024gait, liu2024depression, wang2025ra}, collectively forming WebGait-2M\footnote{We neither process nor redistribute any data; “WebGait-2M” is a convenient name.}.
    This collection comprises over 2.35 million walking sequences, covering a broad spectrum of gait-related challenges (Table~\ref{tab0: WebGait-2m}) and providing a more representative characterization of the real-world distribution of open-source gait data studied by the research community.
\end{itemize}

In extensive experiments, FoundationGait delivers strong and consistent performance across both recognition and healthcare settings. On challenging gait recognition benchmarks—Gait3D~\cite{zheng2022gait} and OU-MVLP~\cite{takemura2018multi}, containing 1,000 in-the-wild and 5,000+ in-the-lab test subjects, respectively—FoundationGait achieves 48.0\% and 64.5\% self-supervised rank-1 accuracy, establishing new milestones for robust cross-condition recognition. In healthcare-oriented tasks, such as scoliosis screening on the Scoliosis1K dataset (1,000+ participants), FoundationGait reaches 97.0\% accuracy after fine-tuning. Detailed evaluations and ablations are provided in Sec.~\ref{sec:experiments}.

Overall, this paper makes three major contributions:
\begin{itemize}
    \item We demonstrate—systematically and for the first time—that vision-based gait pretraining can benefit from empirical scaling trends.
    \item We unify human identification and healthcare tasks within a single, scalable gait foundation model.
    \item Extensive experiments, with and without fine-tuning, verify the strong performance and broad cross-task generalization of \textbf{FoundationGait}.
\end{itemize}

\section{Related Works}
\label{sec:related_works}

Walking patterns, as a fundamental and natural behavior, are widely recognized for their potential to reveal both human identity and health status. 
% ~\cite{niyogi1994analyzing, zhou2024gait, liu2024depression, huang2025generalizable, huangvocabulary,liu2025person}.

\noindent \textbf{Gait Recognition}. 
Gait recognition plays a crucial role in surveillance, particularly in unconstrained and long-distance scenarios. Recent research can be broadly categorized into five main areas~\cite{fan2025opengait}:
(a) Mitigating gait-unrelated factors, such as the influence of RGB-encoded clothing and backgrounds in walking videos. Approaches in this direction include human parsing techniques~\cite{zou2024cross, zheng2023parsing, wang2023gaitparsing}, SMPL models~\cite{wang2025vm,zheng2022gait,wang2025mesh,wang2025combo}, multi-modal learning~\cite{fan2024skeletongait, jin2025exploring}, end-to-end training with inductive biases~\cite{liang2022gaitedge, ye2024biggait, jin2025denoising, ye2025biggergait}, and the use of advanced sensors such as LiDAR~\cite{shen2023lidargait, shen2025lidargait++}.
(b) Enhancing networks with gait-specific priors, such as spatio-temporal and local-global feature learning, combined with attention refinements~\cite{fan2020gaitpart, lin2021gait, wang2023hierarchical, wang2023dygait, ma2023dynamic, peng2024glgait, huang2025h,guo2025gaitcontour}.
(c) Improving the learning process through novel frameworks, such as auto-encoder regression~\cite{guo2023physics}, generative modeling~\cite{jin2025denoising, huang2025origins, ye2023gaitediter, meifield}, causal intervention~\cite{wang2023causal, dou2023gaitgci, xiong2024causality}, and contrastive learning~\cite{wang2025ra, fan2023learning, meng2025mirrorgait}.
(d) Addressing real-world challenges such as occlusion~\cite{xu2023occlusion, huang2024occluded,gupta2025mind,gupta2024you,gupta2025mimicgait}, clothing variation~\cite{li2023depth}, and low-quality data~\cite{wang2024qagait}.
(e) Developing improved loss functions for better optimization~\cite{zhang2021cross, yu2022generalized,su2024open}.

\noindent \textbf{Anomaly Gait Detection}. 
For instance, scoliosis, which affects 0.47\%–5.2\% of adolescents~\cite{konieczny2013scoliosis}, was addressed by Zhou et al.~\cite{zhou2024gait} with the Scoliosis1K dataset, comprising over 1,000 adolescent patients, and the multi-task learning model ScoNet. Similarly, depression impacts 7.1\% of U.S. adults annually~\cite{NIMH2022depression}. Liu et al.~\cite{liu2024depression} released a video-based depression dataset with 100+ participants and evaluated various gait recognition models on it.
Wang et al.~\cite{wang2025ra} developed a gait attribute benchmark involving 15 attributes (e.g., toe-out, head forward, humpback) from 533 subjects and over 120,000 sequences. 
They also created a CLIP-like~\cite{radford2021learning} model for potential use in security and healthcare.

As noted, gait analysis~\cite{wang2025unigait,wang2024hypergait} for both identification and healthcare sometimes benefits from each other. 
For example, Zhou~\cite{zhou2024gait} incorporated identification signals to enhance scoliosis classification, while Wang et al.~\cite{wang2025ra} considered the potential of motion attribute estimation for identification tasks. 
Despite sharing a common focus on human gait, no framework has yet integrated gait recognition and anomaly detection across diverse tasks. 
Our FoundationGait addresses this gap, enabling a scalable and unified solution that leverages the power of large-scale pretraining.

\noindent \textbf{Domain-Specific Foundation Models}. 
Beyond general-purpose LVMs~\cite{chen2021exploring, oquab2023dinov2, kirillov2023segment,xu2025towards}, domain-specific LVMs offer a more practical solution, as they are better suited to understand field data, particularly when it differs significantly from everyday internet frames, such as satellite or industrial images~\cite{guo2024skysense, gu2024AnomalyGPT}. 
In related research, Khirodkar et al.~\cite{khirodkar2024sapiens} introduced a comprehensive foundation model tailored for human-centric vision tasks, including 2D pose estimation, part segmentation, depth estimation, and normal prediction. 
Similarly, Kim et al.~\cite{kim2025sapiensid} developed a foundation model focused on human face~\cite{zhu2025quality,narayan2025facexformer} and body recognition~\cite{su2025hamobe}.
In this paper, FoundationGait takes a step further by focusing on human gait for both recognition and healthcare applications.

\noindent \textbf{Non-Vision-based Gait Foundation Models}.
Beyond vision-based methods, physical signal–driven gait foundation models (e.g., ground reaction forces and joint kinematics)  have recently attracted growing interest.
GaitFM~\cite{yamada2025toward} pre-trains on physics-based synthetic gaits for data-efficient adaptation to dementia-related tasks. GaitDynamics~\cite{tan2026gaitdynamics} employs a diffusion transformer for flexible kinematic modeling and experiment-free prediction of gait modification outcomes. 
PathoFM~\cite{dey2025pathofm} focuses on pathological gait, learning transferable representations from spinal cord injury data through self-supervised objectives.
In contrast, FoundationGait unifies video-based gait tasks, namely, human recognition and healthcare, within a single scalable framework. 
Unless otherwise specified, the remainder of this paper focuses exclusively on the vision-based paradigm.

\section{Method}
\label{sec:method}
Why has the gait research community not yet benefited from empirical scaling trends?
Sec.~\ref{sec:unscale} investigates this issue and provides a detailed analysis, while Sec.~\ref{sec:foundationgait} introduces our proposed solution.
For clarity, experimental settings are presented separately in Sec.~\ref{sec:experiments}.

\subsection{Unscalable Issue in Gait Models}
\label{sec:unscale}
Scalable size is often considered a defining characteristic of foundation models. 
However, existing representative gait models remain relatively small, typically containing fewer than 5M parameters~\cite{chao2019gaitset, fan2023opengait} in their backbone.
To explore the potential of scaling in this domain, we first enlarge the model size.

\begin{figure}[b]
\centering
\includegraphics[width=\linewidth]{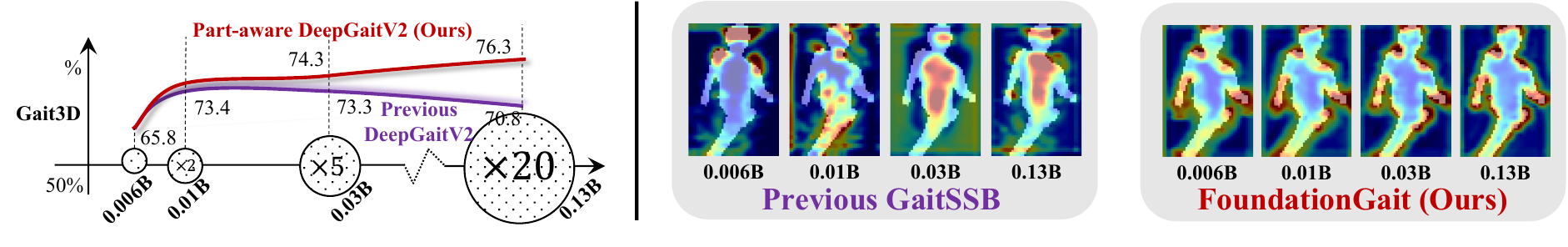}
\captionof{figure}{
\textbf{Left:} \textbf{Unscalable Issue in Supervised DeepGaitV2}~\cite{fan2025opengait}. Our part-aware strategy effectively alleviates this issue.
\textbf{Right:} FoundationGait presents a scale-invariant ability to capture fine-grained full-body details (\textit{e.g.,} hands, contours), whereas GaitSSB prefers brittle shortcuts (\textit{i.e.,} head, chest). The second layer Grad-CAM activation maps for FoundationGait and GaitSSB.
}
\label{fig:unscalable_deepgaitv2}
\end{figure}

\noindent \textbf{Scaling the Supervised DeepGaitV2}. 
As shown in Fig.~\ref{fig:unscalable_deepgaitv2} (left, purple curve), when scaling up DeepGaitV2~\cite{fan2023exploring, fan2025opengait} on the Gait3D dataset~\cite{zheng2022gait}, one of the largest and most challenging gait recognition benchmarks (6K+ seq.), the rank-1 accuracy quickly saturates and even degrades as model size increases.
In contrast, the part-aware strategy proposed by FoundationGait (in Sec.~\ref{sec:foundationgait}) effectively alleviates this issue, exhibiting considerable performance gains as the model scales up (Fig.~\ref{fig:unscalable_deepgaitv2}, left, red curve).

\begin{figure}[t]
\centering
\includegraphics[width=\linewidth]{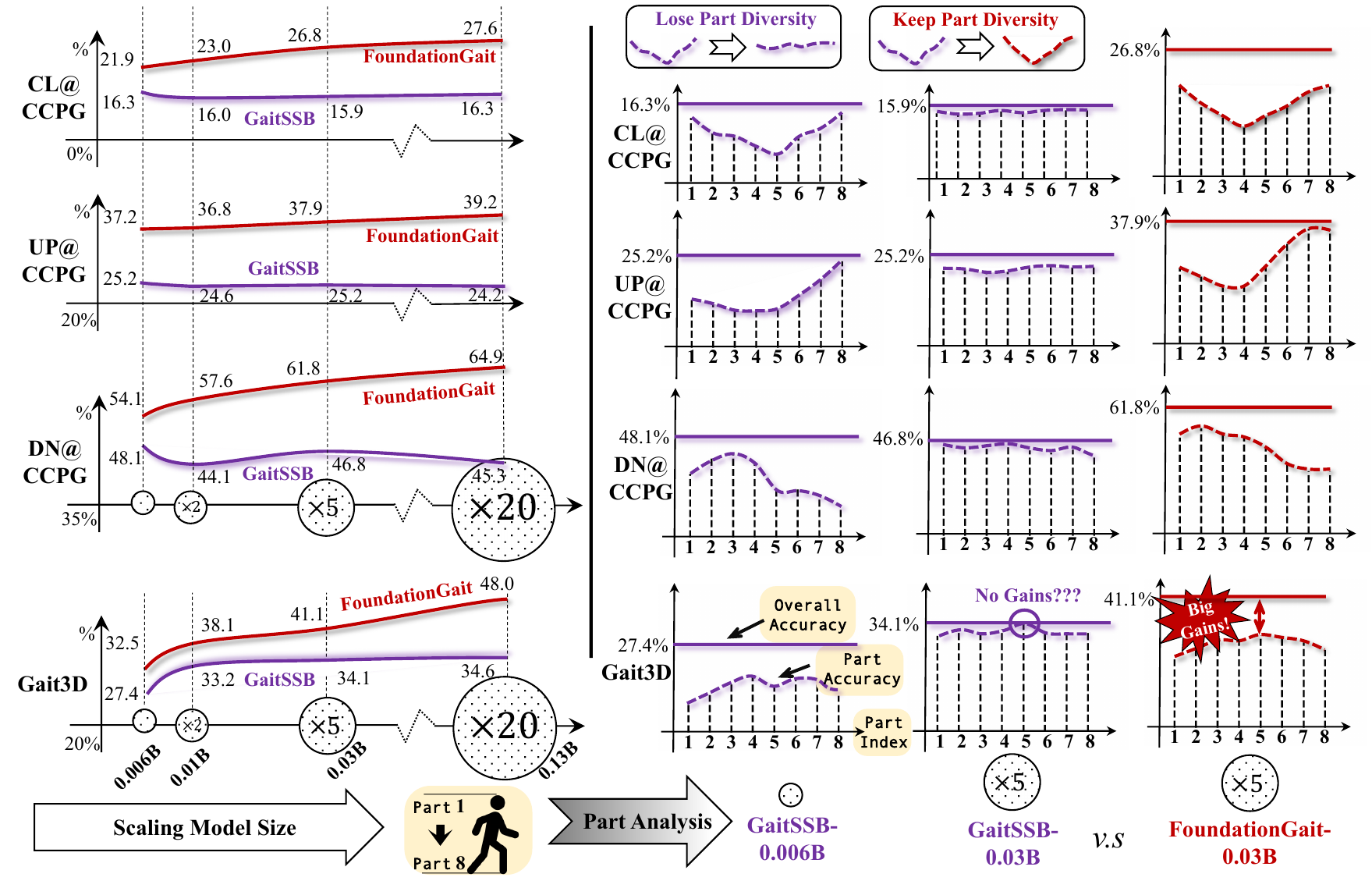}
\captionof{figure}{
\textbf{Unscalable Issue in Self-supervised GaitSSB}~\cite{fan2023learning}.
\textbf{Left:} GaitSSB scales unstably and saturates early, while FoundationGait scales steadily from $2\times$ to $20\times$.
\textbf{Right:} At $5\times$, GaitSSB loses part diversity (Overall $\approx$ Part Accuracy), whereas FoundationGait preserves diverse, complementary parts and achieves larger fusion gains.
All models are pretrained on WebGait-2M.
}
\label{fig:unscalable}
\end{figure}

\noindent \textbf{Scaling the Self-supervised GaitSSB}. 
We further scale up GaitSSB~\cite{fan2023learning}, a representative self-supervised gait pretraining framework, on the large-scale WebGait-2M dataset.
Despite using extensive pretraining data (over 2M+ walking sequences), as shown in Fig.~\ref{fig:unscalable} (left, purple curves), rank-1 accuracy still saturates and even degrades on both CCPG~\cite{li2023depth}, a challenging benchmark with 8K+ sequences spanning diverse clothing variations including full-clothing (CL@CCPG), upper-clothing (UP@CCPG), and lower-clothing (DN@CCPG) changes, and Gait3D~\cite{zheng2022gait}, revealing a fundamental scalability bottleneck in existing approaches.

In contrast, our FoundationGait (red curves in Fig.~\ref{fig:unscalable}, left) largely adheres to the expected scaling trends.
We find that the devil lies in the details, specifically, in the body parts.

The gait community generally agrees that walking patterns require fine-grained descriptions.
Therefore, most popular gait models~\cite{fan2025opengait} adopt a part-based pooling design, dividing the output feature map horizontally into several strips, each assumed to correspond to a specific body region.
\textbf{However, even under this framework, we still observe that}, as shown in Fig.~\ref{fig:unscalable} (right), naively enlarging model size leads to diminished diversity among part features (GaitSSB-0.006B vs. GaitSSB-0.03B, purple curves), resulting in limited fusion gains compared with FoundationGait-0.03B (red curves).

We argue that applying part pooling on an already-computed feature map is fundamentally late for large gait models. 
As networks deepen, part features tend to entangle and become homogeneous, negligible in shallow models but severe at scale.
% The issue cannot be solved by existing part pooling, as it arises from non-independent part encoding.
The issue has been ignored for a long time, and cannot be solved by existing part pooling. 
% We therefore encode parts independently throughout the backbone to prevent cross-part interference and preserve the diversity essential for scaling, forming the core motivation of FoundationGait.
We therefore propose a part-aware encoding framework throughout the backbone to prevent cross-part interference and preserve the diversity necessary for effective scaling, forming the central motivation of FoundationGait.
As further evidenced by Fig.~\ref{fig:unscalable_deepgaitv2} (right), FoundationGait’s part-aware design encourages attention to fine-grained details across the entire body, preventing collapse to a few dominant shortcuts regardless of model scale.

% Intuitively, simply stacking more layers on a well-trained shallow network tends to amplify the most discriminative part features from earlier layers\cite{bender2021dangers, zhang2016understanding}. 
% This overemphasis easily suppresses subtle local cues and reduces feature diversity, ultimately weakening fine-grained gait representation.
% We therefore posit that preserving diversity among part features is the key to scaling up gait models effectively, which serves as the core motivation behind FoundationGait.

\begin{figure}[t]
\centering
\includegraphics[width=1\linewidth, trim=0 60 0 0, clip]{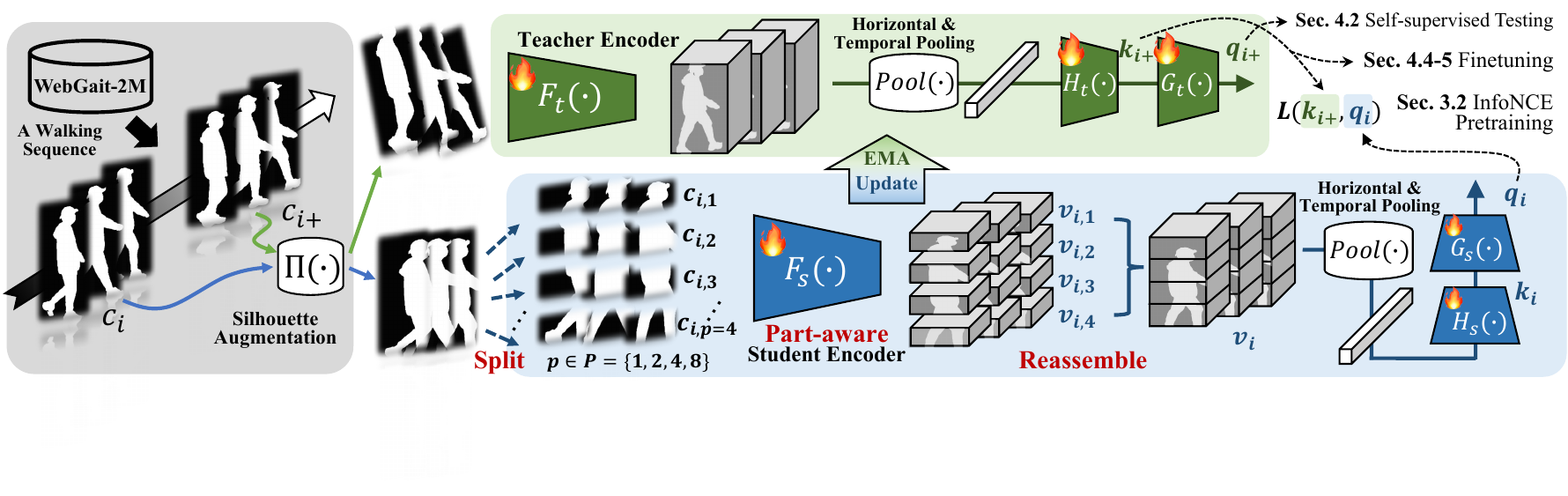}
\caption{  
\textbf{Overview of FoundationGait.}
Two non-overlapping clips are sampled from a sequence and augmented as two views.
The teacher encodes one clip to produce projected feature $k_{i+}$ and predicted feature $q_{i+}$.
The student applies a \textit{Part-aware Strategy} to the other clip: each clip is horizontally split into $p$ parts, encoded independently, and reassembled into a global representation $v_i$ that inherits part-level diversity.
An InfoNCE loss is computed between student predictions $q_i$ and teacher projections $k_{i+}$, with the teacher updated via EMA.
}
\label{fig:pipeline}
\end{figure}

\subsection{FoundationGait}
\label{sec:foundationgait}
% Aiming to alleviate the diversity degradation challenge in large-scale gait models, we introduce FoundationGait, a scalable self-supervised learning framework.

As shown in Fig.~\ref{fig:pipeline}, two non-overlap clips ($c_i$ and $c_{i+}$) are sampled from a sequence as two augmented views of identical walking patterns, where $i\in\{1,...,N\}$.
Each batch contains $2N$ clips forming $N$ positive pairs.
All clips are randomly augmented by $\Pi(\cdot)$.
Following prior self-supervised works~\cite{fan2023learning,grill2020bootstrap}, one clip $c_{i+}$ is encoded by the teacher network to produce a projected feature $k_{i+}$ for downstream tasks and a predicted feature $q_{i+}$ for self-supervised test:
\setlength{\belowdisplayskip}{4pt} \setlength{\belowdisplayshortskip}{4pt}
\setlength{\abovedisplayskip}{4pt} \setlength{\abovedisplayshortskip}{4pt}
\begin{equation}
    % \begin{aligned}
    k_{i+} = H_t(\textit{Pool}(F_t(\Pi(c_{i+})))),\quad
    q_{i+} = G_t(k_{i+}),
    % \end{aligned}
\end{equation}
where $F_t$, $H_t$, and $G_t$ are the teacher’s encoder, projector, and predictor, 
and $\textit{Pool}(\cdot)$ is spatiotemporal aggregation~\cite{fan2025opengait}.
More in \textbf{Supplementary Materials}.

To encourage fine-grained part-level description, we introduce a part-aware training strategy in the student.
Inspired by GaitPart~\cite{fan2020gaitpart}, the other clip $c_{i}$ is first horizontally divided into $p$ parts (with $p=4$ in Fig.~\ref{fig:pipeline} for simplicity), producing $\{c_{i,j}\}^{p}_{j=1}$.
Then, each part is independently fed into the student encoder $F_s$ to obtain  part feature maps $\{v_{i,j}\}^{p}_{j=1}$.
These detail-enriched tensors are then reassembled back into a unified global representation $v_{i}$, that is, 
\begin{equation}
\begin{aligned}
\{v_{i,j}\}_{j=1}^p &= F_s(\mathrm{Split}(\Pi (c_{i}))),\\
v_{i}              &= \mathrm{Concat}(\{v_{i,j}\}_{j=1}^p).
\end{aligned}
\end{equation}
When $p{=}1$, no split is applied, while larger $p$ yields finer local granularity.
To balance local diversity and global consistency, a hyperparameter $P$ (e.g., $p\in P = \{1,2,4,8\}$) is used during training, where $N$ clips are divided into $\text{len}(P)$ subsets, each with its own $p$.
A shift-window operation, similar with SwinViT~\cite{liu2021swin}, is also applied to half of the samples in $\mathrm{Split}(\cdot)$ to enhance local variation, only performed along the height (H) dimension following common practice in gait.
In brief, the global feature map $v_{i}$ is constructed from multiple distinct local ones, \textbf{naturally inheriting their part-level diversity}.

% \textbf{Constructed from these distinct local vectors, the global vector $\mathbf{v_{i+}}$ naturally inherits their part-level diversity.}

% ##################################################################################################
\newlength{\oldfloatsep}        % 声明一个新的长度变量，用来存floatsep的旧值
\newlength{\oldtextfloatsep}    % 声明一个新的长度变量，用来存textfloatsep的旧值
\newlength{\oldintextsep}       % 声明一个新的长度变量，用来存intextsep的旧值
\setlength{\oldfloatsep}{\floatsep}          % 把当前floatsep的值存进去
\setlength{\oldtextfloatsep}{\textfloatsep}  % 把当前textfloatsep的值存进去
\setlength{\oldintextsep}{\intextsep}        % 把当前intextsep的值存进去

\setlength{\floatsep}{0pt}
\setlength{\textfloatsep}{0pt}
\setlength{\intextsep}{0pt}
\begin{figure}[t]
\begin{center}
\begin{minipage}{0.75\linewidth}
\begin{algorithm}[H]
\caption{\small Pseudocode of FoundationGait Framework.}
\label{alg:code}
\algcomment{\fontsize{7.0pt}{0em}\selectfont \texttt{bmm}: batch matrix multiplication; \texttt{mm}: matrix multiplication; \texttt{cat}: concatenation.}
\definecolor{codeblue}{rgb}{0.25,0.5,0.5}
\lstset{
  backgroundcolor=\color{white},
  basicstyle=\fontsize{7.0pt}{7.0pt}\ttfamily\selectfont,
  columns=fullflexible,
  breaklines=true,
  captionpos=b,
  commentstyle=\fontsize{7.0pt}{7.0pt}\color{codeblue},
  keywordstyle=\fontsize{7.0pt}{7.0pt},
  lineskip=-6pt,
}
\begin{lstlisting}[language=python, escapeinside={(*}{*)}]
# F_t,H_t,G_t: teacher encoder, projector, predictor
# F_s,H_s,G_s: student encoder, projector, predictor
# P=[1,2,4,8]: part-aware hyperparameter;
for seq in loader:
    c_t, c_s = sample_two_non_overlapping_clips(seq) 
    c_t, c_s = aug(c_t), aug(c_s)      # two random views
    # ------------ Teacher Branch ------------
    with torch.no_grad():
        v_t = F_t(c_t)                 # (N,C,S,H,W)
        k_t = H_t(Pool(v_t))           # (N,C1,C2)
        q_t = G_t(k_t)                 # (N,C1,C2), Self-supervised feature
    # ------------ Student Branch ------------
    feats = []
    subs = split_batch(c_s, len(P))
    # process each sub-batch with p (*\textcolor[rgb]{0.25,0.5,0.5}{$\in$}*) P
    for clip, p in zip(subs, P):
        parts = Split(clip, p)         # (n,C,S,H,W) -> (n*p,C,S,H//p,W)
        f_part = F_s(parts)
        f_clip = Reassemble(f_part)    # (n*p,C,S,H//p,W) -> (n,C,S,H,W)
        feats.append(f_clip)
    v_s = Concat(feats)                # len(P)*(n,...) -> (N,...)
    q_s = G_s(H_s(Pool(v_s)))         # (N,C1,C2)
    # ------------ Optimization ------------
    loss = InfoNCE(q_s, k_t)
    loss.backward(); optimizer.step()
    T = (F_t, H_t, G_t); S = (F_s, H_s, G_s)
    MomentumUpdate(T, S)               # update teacher by EMA
\end{lstlisting}
\end{algorithm}
\end{minipage}
\end{center}
\end{figure}

\setlength{\floatsep}{\oldfloatsep}          % 还原floatsep
\setlength{\textfloatsep}{\oldtextfloatsep}  % 还原textfloatsep
\setlength{\intextsep}{\oldintextsep}        % 还原intextsep
% ##################################################################################################

Finally, we compute the InfoNCE loss~\cite{gutmann2010noise} between the student’s predictor features $q_{i} = G_s(H_s(\mathrm{Pool}(v_{i})))$ and teacher’s projected output $k_{i+}$, formulated:
\begin{equation}
\begin{aligned}
L(k_{i+}, q_{i}) &= -\log \frac{\exp(\mathrm{sim}(k_{i+}, q_{i})/\tau)}{\sum_{j=1}^{N} \exp(\mathrm{sim}(k_{i+}, q_{j})/\tau)},
\end{aligned}
\end{equation}
where $\mathrm{sim}(\cdot)$ denotes cosine similarity, $\tau=16$ is the temperature, and $j\neq i$ indexes negative sample pairs.
Cross-layer loss follows a common self-supervised paradigm, with asymmetric predictors~\cite{chen2021exploring,grill2020bootstrap} preventing representation collapse.
The pre-training scheme's details are presented in Alg.~\ref{alg:code}. 
For downstream tasks, the teacher model is fine-tuned with its predictor replaced by a new task head; the part-aware training method keeps active for maintaining local diversity.
See \textbf{Supplementary Materials} for details.

\section{Experiment}
\label{sec:experiments}

\subsection{Datasets}
\noindent \textbf{Pretraining Datasets.}
As shown in Tab.~\ref{tab0: WebGait-2m}, we construct WebGait-2M, a large-scale gait database for pretraining, by integrating 12 public datasets spanning both human identification~\cite{yu2006framework, takemura2018multi, song2022casia, zou2024cross, li2023depth, shen2023lidargait, zheng2022gait, zhu2021gait, fan2023learning} and healthcare domains~\cite{zhou2024gait, liu2024depression, wang2025ra}.
Consisting of 2.35M seq. and 0.23B frames, it spans an abundant range of scenes, subjects, motion patterns, clothing styles, viewpoints, and gait attributes, playing a key role in training gait foundation models.
Our FoundationGait targets a universal gait representation generalizable to both recognition and healthcare. 
Thus, we deliberately avoid identity/disease labels, liberating it to learn task-agnostic, bias-free gait patterns.

\noindent \textbf{Recognition Datasets.}
After WebGait-2M pretraining, we evaluate the self-supervised capability on six popular gait datasets, including four in-the-lab datasets (CASIA-B~\cite{yu2006framework}, OUMVLP~\cite{takemura2018multi}, CCPG~\cite{li2023depth} and SUSTech1K~\cite{shen2023lidargait}) and two in-the-wild datasets (Gait3D~\cite{zheng2022gait} and GREW~\cite{zhu2021gait}).
In fine-tuning tasks, experiments are conducted on three representative and challenging datasets: CCPG~\cite{li2023depth}, offering the richest clothing diversity (coats, trousers, backpacks, and full-body outfits); Gait3D~\cite{zheng2022gait}, targeting real-world pedestrian recognition in complex supermarket scenes with frequent occlusions and dynamic motions; and CCGR-MINI~\cite{zou2024cross}, a compact subset of CCGR that preserves its testing difficulty and variations in viewpoints, speeds, and terrains.

\begin{table}[t]
\centering
\renewcommand{\arraystretch}{0.9}
\caption{\textbf{Overview of WebGait-2M.} It is built from twelve public train sets, \textbf{treating each sequence as an individual ID in the self-supervised setting}.
NM, BG, CL, UP, and DN denote normal, bag-, clothing-, ups-, and pants-changing.
}
\resizebox{1\textwidth}{!}{ 

\begin{tabular}{ccccccccc} 
\toprule[2pt]
\multirow{2}{*}{Datasets} & \multirow{2}{*}{Type} & \multirow{2}{*}{Scene} & \multirow{2}{*}{Condition} & \multicolumn{2}{c}{Train Set} & \multicolumn{2}{c}{Test Set}  & \multirow{2}{*}{Frames} \\ 
% \cline{5-8}
\cmidrule(lr){5-6}\cmidrule(lr){7-8}
                          &                       &                        &                            & \#IDs   & \#Seq                                     & \#IDs & \#Seq                                       \\ 
\hline
CASIA-B~\cite{yu2006framework}                   & Recognition           & in-the-lab             & NM, BG, CL                   & 74      & 8,107                                     & 50    & 5,500   & 1,117,083                                    \\ 
CCPG~\cite{li2023depth}                      & Recognition           & in-the-lab             & CL, UP, DN, BG             & 100     & 8,187                                      & 100   & 8,178   & 1,753,431                                   \\ 
CASIA-E~\cite{song2022casia}                  & Recognition           & in-the-lab             & NM, BG, CL                   & 200     & 152,697                                   & 814     & 626,055     & 95,581,249                                     \\ 
OUMVLP~\cite{takemura2018multi}                    & Recognition           & in-the-lab             & NM                         & 5,153   & 133,531                                   & 5,154 & 133,857    & 18,814,160                                \\ 
SUSTech1K~\cite{shen2023lidargait}                 & Recognition           & in-the-lab            & diverse                    & 250     & 6011                                      & 800   & 19,228  & 763,416                                    \\ 
CCGR~\cite{zou2024cross}                      & Recognition           & in-the-lab            & diverse                    & 571     & 908,322                                   & 399     & 672,295   & 150,632,991                                        \\ 
% CCGR-MINI~\cite{zou2024cross}                 & Recognition           & in-the-wild            & diverse                    & -       & -                                         & 399   & 20,377 & -                                     \\ 
Gait3D~\cite{zheng2022gait}                    & Recognition           & in-the-wild            & diverse                    & 3,000   & 18,940                                    & 1,000 & 6,369   & 3,279,239                                    \\ 
GREW~\cite{zhu2021gait}                      & Recognition           & in-the-wild            & diverse                    & 20,000  & 102,887                                   & 6,000 & 24,000  & 13,946,946                                    \\ 
D-Gait~\cite{liu2024depression}                    & Healthcare            & in-the-lab             & Depression                 & 194     & 18,081                                    & 98    & 9,039  & 2,857,343                                     \\ 
RA-GAR~\cite{wang2025ra}                    & Healthcare            & in-the-lab             & Gait Attributes            & 250     & 57,155                                    & 283   & 65,912    & 21,304,199                                  \\ 
Scoliosis1K~\cite{zhou2024gait}              & Healthcare            & in-the-lab             & Scoliosis                  & 745     & 745                                       & 748   & 748   & 447,900                                      \\ 
GaitLU-1M~\cite{fan2023learning}                 & Self-supervised          & in-the-wild            & diverse                    & 943,884 & 943,884                                   & -     & -   & 87,235,933                                        \\ 
\hline
WebGait-2M           & Self-supervised          & mixed                  & diverse                    & 2,358,547  & 2,358,547                                   & -     & -    &  231,675,844                                     \\
\bottomrule[2pt]
\end{tabular}

}
\label{tab0: WebGait-2m}
\end{table}

\begin{table}[t]
\centering
\caption{
(a) \textbf{FoundationGait Configuration.} Mapping denotes the projector ($H$) and predictor ($G$); FLOPs are computed on a $64\times44$ silhouette.
(b) \textbf{Fine-tuning for Recognition.} Batch $(p, q, j)$: $p$ subjects, $q$ sequences each, $j$ frames per sequence.
(c) \textbf{Fine-tuning for Healthcare.}
(d) \textbf{Inference Time.} Averaged over 20 runs after 30 warm-ups on an RTX5090 (FP32) with $30\times64\times44$ input.
}
\resizebox{1.0\columnwidth}{!}{ 
\begin{tabular}{cc}

\begin{tabular}{c}
\begin{tabular}{cccccc} 
\multicolumn{6}{c}{\fontsize{11}{11}\selectfont(a) FoundationGait Configuration.} \\
\toprule[2pt]
\multirow{2}{*}{Model} & \multirow{2}{*}{Layer Depth} & \multicolumn{3}{c}{Parameter Size}  & \multirow{2}{*}{Flops} \\ 
\cline{3-5}
                       &                              & Backbone & Mapping & Total     &                       \\ 
\hline
0.006B              & (1, 1, 1, 1)                 & 5.9M     & 32.6M   & 38.5M    &  1.42G                      \\ 
0.01B               & (1, 4, 4, 1)                 & 11.1M    & 32.6M   & 43.7M           & 2.88G                \\ 
0.03B               & (1, 4, 8, 4)                 & 33.1M    & 32.6M   & 65.7M                 & 6.76G          \\ 
0.13B               & (1, 4, 32, 16)               & 132.3M   & 32.6M   & 164.9M                      & 24.21G    \\
\bottomrule[2pt]
\end{tabular}

\\
\\
\\

\begin{tabular}{cccccc} 
\multicolumn{6}{c}{\fontsize{11}{11}\selectfont(b) Fine-tuning Configuration for Recognition.} \\
\toprule[2pt]
Dataset                    & Model               & Batch Size & Milestones & Total & $P$      \\ 
\hline
\multirow{2}{*}{CCPG~\cite{li2023depth}}      & 0.03B & (16, 32, 30)    & (2K)       & 3K    & [1, 2, 4, 8]  \\ 
                           & 0.13B & (8, 32, 30)     & (2K)       & 3K    & [1, 2, 4, 8]  \\ 
\cline{1-6}
\multirow{2}{*}{Gait3D~\cite{zheng2022gait}}    & 0.03B & (128, 4, 30)    & (3K)       & 4K    & [1, 2]      \\ 
                           & 0.13B & (128, 4, 30)    & (3K, 4K)   & 5K    & [1, 2]      \\ 
\cline{1-6}
\multirow{2}{*}{CCGR-MINI~\cite{zou2024cross}} & 0.03B & (32, 16, 30)    & (3K, 5K)   & 6K    & [1, 2, 4, 8]  \\ 
                           & 0.13B & (32, 16, 30)    & (3K, 5K)   & 6K    & [1, 2, 4, 8]  \\
\bottomrule[2pt]
\end{tabular}
\end{tabular}

&

\renewcommand{\arraystretch}{0.7}
\begin{tabular}{c}
\begin{tabular}{cccccc} 
\multicolumn{6}{c}{\fontsize{11}{11}\selectfont(c) Fine-tuning Configuration for Healthcare.} \\
\toprule[2pt]
Mode                          & Dataset                      & Model   & Milestones & Total & $P$      \\ 
\hline
\multirow{6}{*}{\begin{tabular}[c]{@{}c@{}}Linear \\Probe\end{tabular}} & \multirow{2}{*}{Scoliosis1K~\cite{zhou2024gait}} & 0.03B & (1K, 2K)   & 3K    & [1]        \\
                              &                              & 0.13B & (1K, 2K)   & 3K    & [1]        \\ 
\cline{2-6}
                              & \multirow{2}{*}{D-Gait~\cite{liu2024depression}}      & 0.03B & (1K, 2K)   & 3K    & [1]        \\
                              &                              & 0.13B & (3K, 4K)   & 5K    & [1]        \\ 
\cline{2-6}
                              & \multirow{2}{*}{RA-GAR~\cite{wang2025ra}}      & 0.03B & (1K)       & 2K    & [1]        \\
                              &                              & 0.13B & (1K)       & 2K    & [1]        \\ 
\hline
\multirow{6}{*}{\begin{tabular}[c]{@{}c@{}}Fine- \\tuning\end{tabular}}     & \multirow{2}{*}{Scoliosis1K~\cite{zhou2024gait}} & 0.03B & (1K)       & 2K    & [1, 2, 4, 8]  \\
                              &                              & 0.13B & (1K, 2K)   & 3K    & [1, 2, 4, 8]  \\ 
\cline{2-6}
                              & \multirow{2}{*}{D-Gait~\cite{liu2024depression}}      & 0.03B & (1K)       & 2K    & [1, 2, 4, 8]  \\
                              &                              & 0.13B & (1K, 2K)   & 3K    & [1, 2, 4, 8]  \\ 
\cline{2-6}
                              & \multirow{2}{*}{RA-GAR~\cite{wang2025ra}}      & 0.03B & (1K)       & 2K    & [1, 2, 4, 8]  \\
                              &                              & 0.13B & (1K)       & 2K    & [1, 2, 4, 8]  \\
\bottomrule[2pt]
\end{tabular}

\\
\\

\begin{tabular}{cccccc} 
\multicolumn{6}{c}{\fontsize{11}{11}\selectfont(d) Inference Time Comparison (ms).} \\
\toprule[2pt]
Models & 0.006B & & 0.01B & 0.03B & 0.13B  \\ \midrule
GaitSSB       & 33.36  & & 52.81   & 76.65    & 268.27    \\
FoundationGait & 29.83 & & 49.78   & 82.92    & 273.72    \\
\bottomrule[2pt]
% \multicolumn{5}{l}{Average of 20 runs after 30 warm-up iters on } \\
% \multicolumn{5}{l}{a RTX5090 GPU (FP32) with 30x64×44 input.} \\
\end{tabular}
\end{tabular}
\end{tabular}
}
\label{tab: Model Config}
\label{tab3: finetune implementation}
\label{tab: finetune implementation for healthcare}
\end{table}

\noindent \textbf{Healthcare Datasets.}
In healthcare, three gait-based medical datasets are used: Scoliosis1K~\cite{zhou2024gait}, a large-scale adolescent gait dataset for non-invasive scoliosis screening in real-world settings; D-Gait~\cite{liu2024depression}, designed for depression risk recognition, capturing diverse views and appearances under varying walking conditions; and RA-GAR~\cite{wang2025ra}, a richly annotated dataset with 15 gait attributes, \textit{e.g.}, gender, age, forward head, humpback, and limping.

\begin{table*}[t]
\centering
\caption{
\textbf{Self-supervised Performance Comparison.} FoundationGait is evaluated against self-supervised~\cite{fan2023learning} method on six public benchmarks.
}
\resizebox{1\textwidth}{!}{ 
% \setlength{\tabcolsep}{1pt}
% \Huge
% \LARGE

\begin{tabular}{cccccccccccccc} 
\toprule[2pt]
\multirow{3}{*}{\begin{tabular}[c]{@{}c@{}}Source\\Dataset\end{tabular}} & \multirow{3}{*}{Method} & \multicolumn{12}{c}{Target Dataset} \\ 
\cline{3-14}
                                                                          &                         & \multicolumn{3}{c}{CASIA-B~\cite{yu2006framework}}                                                                  &  OU-MVLP~\cite{takemura2018multi}                              & \multicolumn{2}{c}{Gait3D~\cite{zheng2022gait}}                            & GREW~\cite{zhu2021gait}                                 & \multicolumn{4}{c}{CCPG~\cite{li2023depth}} &  SUSTech1K~\cite{shen2023lidargait}  \\ 
\cmidrule(lr){3-5} \cmidrule(lr){6-6} \cmidrule(lr){7-8} \cmidrule(lr){9-9} \cmidrule(lr){10-13} \cmidrule(lr){14-14}
                                                                          &                         & NM                                   & BG                                   & CL                                   & Rank-1                                  & Rank-1                                  & Rank-5                                  & Rank-1                                  & CL   & UP   & DN   & BG                        & Rank-1        \\ 
\hline
\multirow{3}{*}{\begin{tabular}[c]{@{}c@{}}GaitLU-1M\end{tabular}} & GaitSSB~\cite{fan2023learning}                 & 83.8                                 & \textbf{75.7}                                 & \textbf{28.7}                                 & \textbf{37.2}                                 & 24.8                                 & 38.0                                 & 16.6                                 & 11.3 & 20.0 & \textbf{32.5} & 50.6                      & 50.3       \\ 
                                                                          % & GaitSSB-0.03B                & 84.2                                 & 64.8                                 & 26.0                                 & 39.2                                 & 33.2                                 & 51.2                                 & 24.7                                 & 16.0 & 24.6 & 44.1 & 64.9                      & 41.1       \\ 
                                                                          & GaitSSB-0.13B                & 76.1                                 & 64.9                                 & 23.4                                 & 26.3                                 & 25.1                                 & 39.9 & \textbf{18.7}                                 & 11.1                                 & 20.0 & 31.0 & 49.4                      & 41.3       \\ 
                                                                          % & FoundationGait-0.03B                & 84.2                                 & 64.8                                 & 26.0                                 & 39.2                                 & 33.2                                 & 51.2                                 & 24.7                                 & 16.0 & 24.6 & 44.1 & 64.9                      & 41.1       \\ 
                                                                          & FoundationGait-0.13B                & \textbf{85.0}                                 & 71.2                                 & 25.6                                 & 35.5                                 & \textbf{26.6}                                 & \textbf{40.1}                                 & 18.3                                 & \textbf{12.3} & \textbf{23.0} & 32.1 & \textbf{53.4}                      & \textbf{52.5}       \\ 
\hline
\multirow{5}{*}{\begin{tabular}[c]{@{}c@{}}WebGait-2M\end{tabular}}   & GaitSSB~\cite{fan2023learning}                 & 88.9                                 & 74.8                                 & 33.5                                 & 44.4                                 & 27.4                                 & 43.3                                 & 21.5                                 & 16.3 & 25.2 & 48.1 & 57.6                      & 38.5       \\ 
                                                                          & GaitSSB-0.03B                & 84.2                                 & 64.8                                 & 26.0                                 & 39.2                                 & 33.2                                 & 51.2                                 & 24.7                                 & 16.0 & 24.6 & 44.1 & 64.9                      & 41.1       \\ 
                                                                          & GaitSSB-0.13B                & 82.8                                 & 64.6                                 & 25.9                                 & 38.9                                 & 34.6                                 & 53.2                                 & 23.0                                 & 16.3 & 24.2 & 45.3 & 65.9                      & 36.9       \\ 
                                                                          & FoundationGait-0.03B   & 92.0                                 & 72.9                                 & 39.4                                 & 57.0                                 & 41.1                                 & 59.0                                 & 29.1                                 & 26.8 & 37.9 & 61.8 & 77.1                      & 48.1       \\ 
                                                                          & FoundationGait-0.13B  & \textbf{94.0}                                 & \textbf{75.4}                                 & \textbf{40.2}                                 & \textbf{64.5}                                 & \textbf{48.0}                                 & \textbf{66.5}                                 & \textbf{29.4}                                 & \textbf{27.6} & \textbf{39.2} & \textbf{64.9} & \textbf{80.8}                      & \textbf{49.0}       \\
\bottomrule[2pt]
\end{tabular}
}
\label{tab1: main result}
\end{table*}

\begin{table}[t]
\centering
\caption{
\textbf{Gait Recognition Performance.} Fine-tuning FoundationGait on three challenging public gait datasets: CCPG~\cite{li2023depth}, Gait3D~\cite{zheng2022gait} and CCGR-MINI~\cite{zou2024cross}.
FoundationGait and GaitSSB are pretrained on WebGait-2M.
}
\resizebox{1\columnwidth}{!}{ 
% \setlength{\tabcolsep}{1pt}
% \huge

\begin{tabular}{cc}
\begin{tabular}{c}
\renewcommand{\arraystretch}{0.8}
\begin{tabular}{cccccccc} 
\multicolumn{8}{c}{\fontsize{12}{12}\selectfont(a) Fine-tuning on CCPG~\cite{li2023depth}.} \\
\toprule[2pt]
Input                     & Method            & Venue     & CL   & UP   & DN   & BG   & Mean  \\ 
\hline
\multirow{5}{*}{RGB} & AP3D~\cite{gu2020appearance} & ECCV'20~ & 53.4                                     & 57.3                                     & 69.7                                     & 91.4                                     & 67.8   \\ 
& PSTA~\cite{wang2021pyramid} & ICCV'21 & 42.2                                     & 52.2                                     & 60.3                                     & 84.5                                     & 59.8           \\
& PiT~\cite{zang2022multidirection} & TII’22 & 41.0                                     & 47.6                                     & 64.3                                     & 91.0                                     & 61.0           \\ 
& BigGait~\cite{ye2024biggait} & CVPR'24~ & 82.6  & 85.9  & 87.1  & 93.1  & 87.2   \\ 
& BiggerGait*~\cite{ye2025biggergait} & NeurIPS'25~ & 89.0  & 91.9  & 94.0  & 97.2  & 93.0   \\ 
\hline
\multirow{4}{*}{Ske.} & GaitGraph2~\cite{teepe2021gaitgraph}        & CVPRW'22~ & 5.0  & 5.3  & 5.8  & 6.2  & 5.6   \\ 
                          & Gait-TR~\cite{zhang2023spatial}           & ES'23     & 15.7 & 18.3 & 18.5 & 17.5 & 17.5  \\ 
                          & MSGG~\cite{peng2024learning}              & MTA'23    & 29.0 & 34.5 & 37.1 & 33.3 & 33.5  \\ 
                          & SkeletonGait~\cite{fan2024skeletongait}      & AAAI'24   & 40.4 & 48.5 & 53.0 & 61.7 & 50.9  \\ 
\hline
\multirow{8}{*}{Sil.}    & GaitSet~\cite{chao2019gaitset}           & AAAI'19   & 60.2 & 65.2 & 65.1 & 68.5 & 64.8  \\ 
                          & GaitPart~\cite{fan2020gaitpart}          & CVPR'20   & 64.3 & 67.8 & 68.6 & 71.7 & 68.1  \\ 
                          & OGBase~\cite{li2023depth}            & CVPR'23   & 52.1 & 57.3 & 60.1 & 63.3 & 58.2  \\ 
                          & GaitBase~\cite{fan2023opengait}          & CVPR'23   & 71.6 & 75.0 & 76.8 & 78.6 & 75.5  \\ 
                          & DeepGaitV2~\cite{fan2025opengait}        & TPAMI'25  & 78.6 & 84.8 & 80.7 & 89.2 & 83.3  \\ 
                          & GaitSSB-0.13B~\cite{fan2023learning}    & TPAMI'23      & 50.8     & 56.3 & 55.2 & 61.5 & 56.0 \\ 
\cline{2-8}
                          & FoundationGait-0.03B    & ours      & 80.0     & 86.0 & 83.7 & 90.8 & 85.1 \\ 
                          & FoundationGait-0.13B & ours      &   \textbf{84.0}  &   \textbf{87.5}   &   \textbf{85.5}   &   \textbf{91.7}   & \textbf{87.2}   \\
\bottomrule[2pt]
\end{tabular}

\\
\\

\renewcommand{\arraystretch}{0.9}
\begin{tabular}{ccccc} 
\multicolumn{5}{c}{\fontsize{12}{12}\selectfont(b) Fine-tuning on CCGR-MINI~\cite{zou2024cross}.} \\
\toprule[2pt]
Method            & Venue    & Rank-1 & mAP   & mINP   \\ 
\hline
GaitSet~\cite{chao2019gaitset}           & AAAI'19  & 13.77  & 15.39 & 5.75   \\ 
% \hline
GaitPart~\cite{fan2020gaitpart}          & CVPR'20  & 8.02   & 10.12 & 3.52   \\ 
% \hline
GaitGL~\cite{lin2022gaitgl}            & ICCV'21  & 17.51  & 18.12 & 6.85   \\ 
% \hline
GaitBase~\cite{fan2023opengait}          & CVPR'23  & 26.99  & 24.89 & 9.72   \\ 
% \hline
DeepGaitV2~\cite{fan2025opengait}        & TPAMI'25 & 39.37  & 36.01 & 16.77  \\ 
GaitSSB-0.13B~\cite{fan2023learning}        & TPAMI'23 & 26.06  & 26.69 & 15.66  \\ 
\hline
FoundationGait-0.03B    & ours     & 45.66      & 40.97     & 26.36      \\ 
% \hline
FoundationGait-0.13B & ours     & \textbf{53.25}      & \textbf{48.00}     & \textbf{33.29}      \\
\bottomrule[2pt]
\end{tabular}

\end{tabular}
&
\begin{tabular}{c}
\begin{tabular}{ccccc} 
\multicolumn{5}{c}{\fontsize{12}{12}\selectfont(c) Fine-tuning on Gait3D~\cite{zheng2022gait}.} \\
\toprule[2pt]
Method            & Venue    & Rank-1 & Rank-5 & mAP   \\ 
\hline
GaitSet~\cite{chao2019gaitset}           & AAAI'19  & 36.7   & 58.3   & 30.0  \\ 
% \hline
GaitPart~\cite{fan2020gaitpart}          & CVPR'20  & 28.2   & 47.6   & 47.6  \\ 
% \hline
GaitGL~\cite{lin2022gaitgl}            & ICCV'21  & 29.7   & 48.5   & 22.3  \\ 
% \hline
GaitContour~\cite{guo2025gaitcontour}        & WACV'25  & 25.3   & 41.3      & -  \\ 
SMPLGait~\cite{zheng2022gait}          & CVPR'22  & 46.3   & 64.5   & 37.2  \\ 
% \hline
MTSGait~\cite{zheng2022gait2}           & MM'22    & 48.7   & 67.1   & 37.6  \\ 
% \hline
QAGait~\cite{wang2024qagait}            & AAAI'24  & 67.0   & 81.5   & 56.5  \\ 
% \hline
GaitBase~\cite{fan2023opengait}          & CVPR'23  & 64.6   & -      & -     \\ 
% \hline
DANet~\cite{ma2023dynamic}             & CVPR'23  & 48.0   & 69.7   & -     \\ 
% \hline
GaitGCI~\cite{dou2023gaitgci}           & CVPR'23  & 50.3   & 68.5   & 39.5  \\ 
% \hline
DyGait~\cite{wang2023dygait}            & ICCV'23  & 66.3   & 80.8   & 56.4  \\ 
% \hline
HSTL~\cite{wang2023hierarchical}              & ICCV'23  & 61.3   & 76.3   & 55.5  \\ 
% \hline
VPNet~\cite{ma2024learning}             & CVPR'24  & 75.4   & 87.1   & -     \\ 
% \hline
DeepGaitV2~\cite{fan2025opengait}        & TPAMI'25 & 74.4   & 88.0   & 65.8  \\ 
% \hline
CLTD~\cite{xiong2024causality}              & ECCV'24  & 69.7   & 85.2   & -     \\ 
% \hline
GaitMoE~\cite{huang2024occluded}           & ECCV'24  & 73.7   & -      & 66.2  \\ 
% \hline
Free Lunch~\cite{wang2024free}        & ECCV'24  & 70.1   & -      & 61.9  \\ 
VM-Gait~\cite{wang2025vm}        & WACV'25  & 75.4   & 87.5      & 66.4  \\ 
Mesh-Gait~\cite{wang2025mesh}        & ArXiv'25  & 75.0   & 88.0      & 66.1  \\ 
GaitSSB-0.13B~\cite{fan2023learning}        & TPAMI'23  & 53.7   & 75.5      & 48.7  \\ 
\hline
FoundationGait-0.03B    & ours     & 77.7      & \textbf{89.6}      & 71.1     \\ 
% \hline
FoundationGait-0.13B & ours     & \textbf{79.3}      & 89.5      & \textbf{74.0}     \\
\bottomrule[2pt]
\end{tabular}
\end{tabular}

\end{tabular}

}
\label{tab: CCGR-MINI}
\label{tab: Gait3D}
\label{tab: CCPG}
\end{table}

\subsection{Implementation Details}
\label{sec: Implementation Details}
\noindent \textbf{Protocols}. 
Silhouettes are size-normalized following~\cite{fan2023opengait}, resized to $64\times44$, and augmented following~\cite{fan2023learning}.
All experiments strictly follow the official protocols. 
% Gait evaluation protocols and rank-1 accuracy are reported.
% Gait evaluation protocols are reported, using rank-1 accuracy as the main metric. 

\noindent \textbf{Network.} FoundationGait uses the popular DeepGaitV2's backbone~\cite{fan2025opengait} as its encoder, scaled up by stacking layers.
See Tab.~\ref{tab: Model Config} (a) and (d) for exact model size, computation costs and inference time. 

\noindent \textbf{Pretraining.}
SGD is used with an initial learning rate of $0.05$, momentum $0.9$, and weight decay $5\times10^{-4}$.
The 0.13B model is trained for 80K iterations with a cosine annealing scheduler ($T_{max}=80$K, $\eta_{min}=1\times10^{-5}$), while the others are trained for 40K iterations using MultiStepLR (20K, 30K; $\gamma=0.1$).
The EMA momentum rises from $0.99$ to $1.0$ via a cosine schedule.
Batch size is 512 with 16 frames per clip.
In the part-aware student, $P=[1,2,4,8]$.

\noindent \textbf{Sampling.}
To balance data diversity, each batch samples from the 12 datasets with softmax-normalized probabilities proportional to $\log(\text{size})/3.0$, while \textit{GaitLU}-1M is downweighted by $10\times$ due to its limited viewpoint diversity~\cite{ma2023dynamic}.
The 0.13B model takes roughly 178 hours on 16$\times$ 32GB V100 GPUs.
Fine-tuning 0.13B on CCPG takes 9.5 h and on Scoliosis1K 2.2 h, on 8$\times$24GB RTX A5000.

\subsection{Self-supervised Performance}
For brevity, only the two largest FoundationGait versions (0.03B and 0.13B) are compared against representative self-supervised~\cite{fan2023learning} methods across six widely used benchmarks in Tab.~\ref{tab1: main result}. 
Note that self-supervised gait pretraining remains in its infancy, with GaitSSB~\cite{fan2023learning} being the only reproducible baseline.

First, when pre-training on GaitLU-1M~\cite{fan2023learning}, scaling GaitSSB~\cite{fan2023learning} from 0.006B to 0.13B yields significant performance drops across most datasets — $-7.7$\% on CASIA-B (NM), $-10.8$\% (BG), $-10.9$\% on OU-MVLP, and $-9.0$\% on SUSTech1K. 
By contrast, FoundationGait-0.13B remains competitive with the small GaitSSB, owing to its stronger retention of fine-grained details. 
However, a key observation is that neither scaled-up model benefits meaningfully from GaitLU-1M beyond the small baseline.
We attribute this to the dataset's limited diversity: sourced from handheld internet videos, it offers constrained viewpoint variation and is prone to causing overfitting in larger models~\cite{ma2023dynamic}.
Therefore, we turn to the larger and more diverse WebGait-2M for scaling experiments.

Second, when pre-training on WebGait-2M, the scaling trend begins to pay off.
Compared with the SoTA self-supervised method GaitSSB~\cite{fan2023learning}, our 0.13B model achieves notable gains: +20.6\% on the challenging in-the-wild Gait3D ~\cite{zheng2022gait}, +20.1\% on the largest indoor OU-MVLP~\cite{takemura2018multi}, and +11.3\% on the CCPG (CL)~\cite{li2023depth}.
However, even with richer pretraining data, scaling up GaitSSB to 0.13B yields limited gains and even regressions (\textit{e.g.}, –5.5\% on OU-MVLP).
This demonstrates that simply enlarging model size and pretraining data is insufficient — \textbf{method-level innovations, as in FoundationGait, are essential}.

\subsection{Fine-tuning on Gait Recognition}
\label{Sec: Finetuning on Recognition}
\noindent \textbf{Implementation.}
Following GaitSSB~\cite{fan2023learning}, we adopt layer-wise fine-tuning.
Only the last two backbone blocks are fine-tuned (lr = 0.001, 0.005), while the projector and head use 0.01 and 0.1.
SGD (lr = 0.1, momentum = 0.9) with a MultiStepLR scheduler is used.
More details are in Tab.~\ref{tab3: finetune implementation} (b).

\noindent \textbf{CCPG.}
In Tab.~\ref{tab: CCPG} (a), our 0.13B model delivers notable improvements in this challenging clothing-changing scene: +5.4\% in full-changing (CL), +2.7\% in ups-changing (UP), +4.8\% in pants-changing (DN), and +3.9\% on average.
Despite relying solely on informative-sparse silhouettes as input, our FoundationGait-0.13B achieves performance comparable to BigGait@CVPR'24—which leverages RGB inputs and large vision model features (84.0\% \textit{vs.} 82.6\% on CL; 87.2\% \textit{vs.} 87.2\% on Mean).
Though a gap remains with the SoTA RGB foundation model, BiggerGait~\cite{ye2025biggergait}, we consider that advances in Silhouette-based methods often inspire RGB ones (\textit{e.g.}, BiggerGait builds upon GaitBase).

\noindent \textbf{CCGR-MINI.}
In Tab.~\ref{tab: CCGR-MINI} (b), even in this complex covariate setting, existing gait methods have long struggled (Rank-1 $<40\%$), but FoundationGait offers a significant breakthrough, \textit{i.e.}, 45.66\% / 53.25\% on Rank-1 for our 0.03B / 0.13B.

\noindent \textbf{Gait3D.}
In Tab.~\ref{tab: Gait3D} (c), FoundationGait establishes a new SoTA on this challenging real world scene.
Our 0.13B model achieves 79.3\% Rank-1 and 74.0\% mAP, significantly outperforming the previous best methods, VPNet~\cite{ma2024learning} (+3.9\% in Rank-1) and GaitMoE~\cite{huang2024occluded} (+7.8\% in mAP).

% In the challenging CCGR-MINI~\cite{zou2024cross}, including 53 combinations of diverse conditions, silhouette-based gait methods have long struggled to perform well on it.
% Our FoundationGait achieves a significant breakthrough, offering a promising direction for such complex gait scenes.
% As shown in Tab.~\ref{tab: CCGR-MINI}, FoundationGait-60M surpasses the previous state-of-the-art DeepGaitV2 by +6.29\% in Rank-1 and +4.96\% in mAP.
% The larger FoundationGait-0.2B further achieves outstanding improvements of +13.88\% in Rank-1, +11.99\% in mAP, and +16.52\% in mINP.
% These substantial gains highlight the strong potential of FoundationGait in handling highly complex and challenging real-world scenarios.

\subsection{Fine-tuning on Healthcare Tasks}
\noindent \textbf{Implementation.}
Two settings (linear probing and fine-tuning) are evaluated.
For linear probing, only the classification head is trained (lr = 0.1).
For fine-tuning, the model is initialized from the linear probe stage and trained with layer-wise learning rates.
The 0.03B model uses learning rates of $1\times10^{-3}$, $1\times10^{-2}$, and $1\times10^{-1}$ for the backbone, projector, and head, respectively. 
Other settings follow Sec.~\ref{Sec: Finetuning on Recognition}.
Batch size is (8, 8, 30), following the setting in Tab.~\ref{tab3: finetune implementation} (b).
More details are in Tab.~\ref{tab: finetune implementation for healthcare} (c).
% The evaluation protocol and code are reproduced according to~\cite{zhou2024gait,liu2024depression,wang2025ra}.

\noindent \textbf{Linear Probe}. 
As demonstrated in Tables~\ref{tab: Scoliosis1K} (a-c), the features extracted by FoundationGait exhibit strong linear separability across a variety of gait healthcare tasks. Specifically, for gait attribute estimation~\cite{wang2025ra} (+2.2\% in F1 compared to CLIP-GAR~\cite{wang2025ra}), Foundation-0.13B performs on par with, or even surpasses, previous SoTA methods that require full training. 
These results underscore the model's superior ability to capture abundant and informative gait characteristics that are \textbf{already highly linearly separable in its feature space}, contributing to its strong generalizability across vision-based gait tasks.

\noindent \textbf{Fine-tuning}. 
After fine-tuning, FoundationGait demonstrates significant performance gains, with +6.8\% in F1 on the Scoliosis1K~\cite{zhou2024gait} and +7.3\% in F1 on the D-Gait~\cite{liu2024depression} datasets. 
However, FoundationGait-0.13B tends to be overly optimistic in predicting positive cases, leading to high recall but lower accuracy compared to the 0.03B version on these class-imbalanced benchmarks. 
This suggests that the model size may influence the tendency to classify more samples as positive in long-tailed distributions. 
For the gait attribute estimation task (RA-GAR~\cite{wang2025ra}), where labels are already well separable through linear probing of the pretrained FoundationGait, further fine-tuning yields limited gains.

In summary, FoundationGait demonstrates remarkable transferability across multiple gait healthcare tasks, achieving strong performance with simple linear probing or fine-tuning without task-specific designs. We believe that more refined strategies, such as LoRA~\cite{ding2024lora} and Adapter~\cite{chen2024conv}, could further unlock its potential for healthcare gait analysis.

\begin{table}[t]
\centering
\caption{
\textbf{Healthcare Performance.} Linear-probe and fine-tuning results of FoundationGait on scoliosis screening~\cite{zhou2024gait}, depression prediction~\cite{liu2024depression}, and gait attribute~\cite{wang2025ra}.
FoundationGait and GaitSSB are pretrained on WebGait-2M.
}
\resizebox{1\columnwidth}{!}{ 

\begin{tabular}{cc}

\begin{tabular}{c}

\renewcommand{\arraystretch}{0.85}
\begin{tabular}{cccccc} 
\multicolumn{6}{c}{\fontsize{12}{12}\selectfont(a) Evaluation on Scoliosis1K~\cite{zhou2024gait}.} \\
\toprule[2pt]
Mode                                                                    & Method              & Acc.                    & Precision              & Recall        & F1                      \\ 
\hline
% \multirow{2}{*}{\small Supervised}
\multirow{2}{*}{\begin{tabular}[c]{@{}c@{}}Full- \\training\end{tabular}} & ScoNet~\cite{zhou2024gait}              & 93.3                   & 83.4                   & 92.5          & 86.4                    \\
                                                                        & ScoNet-MT~\cite{zhou2024gait}           & 95.2                   & 80.2                   & 96.6          & 84.9                    \\ 
\hline
\multirow{3}{*}{\begin{tabular}[c]{@{}c@{}}Linear \\Probe\end{tabular}}
                                                                        & GaitSSB-0.13B & 72.5                   & 56.7                   & 69.8          & 58.9                    \\
                                                                        & FoundationGait-0.03B & 71.1                   & 59.7                   & 75.3          & 56.7                    \\
                                                                        & FoundationGait-0.13B & 73.5                   & 61.0                   & 68.5          & 56.9                    \\ 
\hline
\multirow{3}{*}{\begin{tabular}[c]{@{}c@{}}Fine-\\tuning\end{tabular}}
                                                                        & GaitSSB-0.13B & 93.7                   & 91.0                   & 74.6          & 79.5                    \\
                                                                        & FoundationGait-0.03B & \textbf{\textbf{97.3}} & \textbf{\textbf{95.4}} & 88.9          & \textbf{\textbf{91.7}}  \\
                                                                        & FoundationGait-0.13B & 96.0                   & 80.1                   & \textbf{97.5} & 85.1                    \\
\bottomrule[2pt]
\end{tabular}

\\
\\

\renewcommand{\arraystretch}{0.85}
\begin{tabular}{cccccc} 
\multicolumn{6}{c}{\fontsize{12}{12}\selectfont(b) Evaluation on D-Gait~\cite{liu2024depression}.} \\
\toprule[2pt]
Mode                                                                    & Method              & Acc.                   & Precision              & Recall        & F1            \\ 
\hline                                        
\multirow{3}{*}{\begin{tabular}[c]{@{}c@{}}Full- \\training\end{tabular}} & GaitBase~\cite{fan2023opengait}            & -                      & 51.7                   & 60.4          & 55.7           \\
                                                                        & GaitSet~\cite{chao2019gaitset}             & -                      & 53.8                   & 65.7          & 59.2           \\
                                                                        & GaitGL~\cite{lin2022gaitgl}              & -                      & 57.5                   & 36.8          & 44.9           \\ 
\hline
\multirow{3}{*}{\begin{tabular}[c]{@{}c@{}}Linear \\Probe\end{tabular}}
                                                                        & GaitSSB-0.13B & 65.0                   & 56.9                   & 57.3          & 57.1                    \\
                                                                        & FoundationGait-0.03B & 66.3                   & 57.7                   & 57.8          & 57.8           \\
                                                                        & FoundationGait-0.13B & 66.0                   & 57.7                   & 58.0          & 57.8           \\ 
\hline
\multirow{3}{*}{\begin{tabular}[c]{@{}c@{}}Fine-\\tuning\end{tabular}}
                                                                        & GaitSSB-0.13B & 72.9                   & 64.9                   & 62.9          & 63.6                    \\
                                                                        & FoundationGait-0.03B & \textbf{\textbf{75.1}} & \textbf{\textbf{67.9}} & 65.1          & 66.1           \\
                                                                        & FoundationGait-0.13B & 72.2                   & 66.0                   & \textbf{67.4} & \textbf{66.5}  \\
\bottomrule[2pt]
\end{tabular}

\end{tabular}

&

\begin{tabular}{c}

\begin{tabular}{ccccccc} 
\multicolumn{7}{c}{\fontsize{12}{12}\selectfont(c) Evaluation on RA-GAR~\cite{wang2025ra}.} \\
\toprule[2pt]
\multirow{2}{*}{Mode}                       & \multirow{2}{*}{Method} & \multicolumn{4}{c}{Instance-based}                            & Attr.-based    \\ 
\cmidrule(l){3-6}\cmidrule(lr){7-7}
                                              &                        & Acc.          & Prec.         & Recall        & F1           & mA             \\ 
\hline
% \multirow{5}{*}{Supervised}
\multirow{7}{*}{\begin{tabular}[c]{@{}c@{}}Full- \\training\end{tabular}} &   GaitSet~\cite{chao2019gaitset}              & 84.4          & 75.7          & 66.1          & 70.2          & 64.3           \\
& GaitGL~\cite{lin2022gaitgl} & 85.0          & 78.0          & 64.9          & 70.4          & 62.6           \\
& GaitTR~\cite{zhang2023spatial} & 83.6          & 73.6          & 65.0          & 68.7          & 62.3           \\
                                        & GPGait~\cite{fu2023gpgait}                & 84.3          & 75.0          & 66.3          & 70.0          & 63.9           \\
                                       & GAR-Net~\cite{song2023gait}                & 84.3          & 77.4          & 74.7          & 73.2          & 62.2           \\
                                    & DeepGaitV2~\cite{fan2025opengait}               & 85.0          & 76.6          & 66.7          & 70.9          & 61.7           \\
                                      & CLIP-GAR~\cite{wang2025ra}                & 84.3          & 77.4          & 74.7          & 76.0          & 65.6           \\ 
\hline
\multirow{3}{*}{\begin{tabular}[c]{@{}c@{}}Linear \\Probe\end{tabular}}         & GaitSSB-0.13B                   & 84.0          & 77.1          & 74.2          & 75.5          & 61.8           \\
         & FoundationGait-0.03B                   & 85.2          & 78.6          & 76.4          & 77.4          & 64.3           \\
         & FoundationGait-0.13B                   & \textbf{85.7} & \textbf{79.3} & \textbf{77.3} & \textbf{78.2} & 65.6           \\
\hline
\multirow{3}{*}{\begin{tabular}[c]{@{}c@{}}Fine-\\tuning\end{tabular}}         & GaitSSB-0.13B                   & 84.7          & 78.7          & 74.4          & 76.4          & 62.9           \\
& FoundationGait-0.03B                   & 84.8          & 78.0          & 75.9          & 76.9          & 65.1           \\
 & FoundationGait-0.13B                   & 84.7          & 78.1          & 75.5          & 76.7          & \textbf{65.7}  \\
\bottomrule[2pt]
\end{tabular}
\end{tabular}

\end{tabular}

}
\label{tab: Scoliosis1K}
\label{tab: D-Gait}
\label{tab: RA-GAR}
\end{table}

\subsection{Ablation Study}
\label{sec: Ablation Study}
We systematically evaluate:
(a) the effectiveness of part-aware student and its hyperparameter $P$, 
(b) part-aware extensions in supervised models, 
(c) cross-domain generalization, and (d) cross-modality generalization of FoundationGait.
See \textbf{Supplementary Materials} for experiment details.

\noindent \textbf{Part-aware Effectiveness.}
Compared to cropping~\cite{zhang2025procrop} or masking silhouettes (Tab.~\ref{tab: part-aware ablation} (a)), our part-aware student cleverly preserves all details, achieving stronger self-supervised results.
Expanding $P$ from $[1]$ to $[1,2,4,8]$ provides clear gains on indoor CASIA-B~\cite{yu2006framework} and OU-MVLP~\cite{takemura2018multi}.

% \noindent \textbf{Part-aware Effectiveness.}
% % Instead of throwing data away (\textit{e.g.}, Crop and Mask in Tab.~\ref{tab: part-aware ablation}), 
% Compared to cropping~\cite{zhang2025procrop} or masking silhouettes (Tab.~\ref{tab: part-aware ablation}), our part-aware student cleverly preserves all details, achieving stronger self-supervised results.
% Expanding $P$ from $[1]$ to $[1,2,4,8]$ causes a minor drop on the in-the-wild Gait3D~\cite{zheng2022gait} but clear gains on indoor CASIA-B~\cite{yu2006framework} and OU-MVLP~\cite{takemura2018multi}.
% This likely reflects a stronger focus on local cues and is considered acceptable (+12.7\% on OU\text{-}MVLP \textit{vs.} -1.0\% on Gait3D), so $P=[1,2,4,8]$ is adopted.

\noindent \textbf{Extension in Supervised Models.}
Beyond the self-supervised setting, can part-aware training be applied to large supervised models?
Table~\ref{tab: part-awared on Supervised Method} (b) demonstrates that our part-aware training significantly benefits DeepGaitV2 with large model sizes. For example, the 0.13B version shows improvements of +5.5\% on Gait3D and up to +20.2\% on CASIA-B.
This underscores the importance of preserving part diversity in large gait models, both in supervised and self-supervised settings, and  our part-aware training offers a flexible and promising solution. 

% \noindent \textbf{ViT Replacement.}
% % Although transformers are widely used as the backbone of modern foundation models, their performance on gait tasks still lags behind that of CNNs.
% % To examine this within the FoundationGait framework, we compare CNN- and ViT-based variants under similar settings.
% Following SwinGait~\cite{fan2025opengait}, we replace the last two CNN blocks (of sizes $(8, 4)$) in FoundationGait-0.03B with a 12-layer ViT block, resulting in a noticeable performance drop (see Table~\ref{tab: ViT} (c)).
% The reasons for this decline remain unclear and are not specific to FoundationGait. In fact, CNN-based architectures continue to dominate vision-based gait methods~\cite{fan2025opengait}.
% % Given that most modern foundation models, particularly in the natural language~\cite{weiopen} domain, are transformer-based~\cite{vaswani2017attention}, exploring ViT-based gait foundation models remains a challenging but crucial avenue for future research.
% % This suggests that CNNs still outperform ViTs in the FoundationGait framework.
% % Developing ViT-based gait foundation models remains a worthwhile direction.

\noindent \textbf{Cross-domain Generalization}. Results on DroneGait further validate FoundationGait's strong generalization to unseen UAV-view domains (Table~\ref{tab: ViT} (c)), consistently outperforming GaitSSB across both mid- and high-pitch views.

\noindent \textbf{Cross-modality Generalization}. 
Although FoundationGait has never been pretrained on the gait modality of human parsing, we found that it performs well in both parsing-only and multi-modal settings after fine-tuning (Tab.~\ref{tab: cross modality} (d)). 
This highlights FoundationGait's ability to transform raw pixel-based shapes, across diverse data types, into fine-grained semantic features.

\begin{table}[t]
\centering
\caption{\textbf{Ablation Study.} 
(a) Exploring the effectiveness of part-aware training and its P , using FoundationGait-0.13B for pretraining. 
(b) Extending part-aware training to DeepGaitV2, where w/ denotes the use of part-aware training.
(c) Evaluating FoundationGait on the unseen domain datasets~\cite{li2023gait}.
(d) Fine-tuning FoundationGait on Gait3D with taking the unseen human parsing as input.
}
\resizebox{1\columnwidth}{!}{

\begin{tabular}{cc}

\begin{tabular}{c}

\renewcommand{\arraystretch}{0.9}
\begin{tabular}{ccccccc} 
\multicolumn{7}{c}{\fontsize{11}{11}\selectfont(a) Ablation of part-aware training and $P$.} \\
\toprule[2pt]
\multirow{2}{*}{Method} & \multirow{2}{*}{$P$} & \multicolumn{3}{c}{CASIA-B~\cite{yu2006framework}} & \multicolumn{2}{c}{Gait3D~\cite{zheng2022gait}}  \\ 
\cmidrule(lr){3-5}\cmidrule(lr){6-7}
                        &                              & NM   & BG   & CL                                  & Rank-1 & mAP \\ 
\hline
Crop              & - & 70.8                             & 59.3    & 27.9    & 17.2                                   & 12.3 \\ 
Mask             & - & 87.6                            & 68.1   & 36.2    & 38.1                                   & 29.4 \\
\hline
\multirow{3}{*}{\begin{tabular}[c]{@{}c@{}}Part- \\aware\end{tabular}} & [1]                          & 86.6 & 67.0 & 28.7                                & 39.6 & 30.2 \\ 
                      & [1, 4]                       & 92.2    & 73.8    & 37.7                                   & 45.3 & 60.3 \\ 
                     & [1, 2, 4, 8]                 & \textbf{94.0} & \textbf{75.4} & \textbf{40.2}                                & \textbf{48.0} & \textbf{66.5} \\ 
\bottomrule[2pt]
\end{tabular}

\\
\\

\begin{tabular}{ccccccc} 
\multicolumn{7}{c}{\fontsize{12}{12}\selectfont(b) Extending to supervised methods.} \\
\toprule[2pt]
\multirow{2}{*}{Model}      & \multicolumn{1}{c}{\multirow{2}{*}{w/}} & \multicolumn{3}{c}{CASIA-B~\cite{yu2006framework}}                                                                                       & \multicolumn{2}{c}{Gait3D~\cite{zheng2022gait}}                                           \\ 
\cmidrule(lr){3-5}\cmidrule(lr){6-7}
                             & \multicolumn{1}{c}{}                     & NM                                          & BG                                          & CL                                          & Rank-1                                      & mAP                                          \\ 
\hline
\multirow{2}{*}{\begin{tabular}[c]{@{}c@{}}DeepGaitV2 \\-0.03B\end{tabular}}   & $\times$                                                              & 84.5                                        & 72.3                                        & 46.5                                        & 73.3                                        & 65.6                                         \\ 
% \cline{2-7}
                             & \checkmark                                                              & \textbf{92.4} & \textbf{87.0} & \textbf{66.2} & \textbf{74.3} & \textbf{67.3}  \\ 
\hline
\multirow{2}{*}{\begin{tabular}[c]{@{}c@{}}DeepGaitV2 \\-0.13B\end{tabular}} & $\times$                                                              & 79.2                                        & 68.0                                        & 42.2                                        & 70.8                                        & 64.0                                         \\ 
% \cline{2-7}
                             & \checkmark                                                              & \textbf{90.0} & \textbf{81.4} & \textbf{62.4} & \textbf{76.3} & \textbf{68.7}  \\
\bottomrule[2pt]
\end{tabular}

\end{tabular}
&
\begin{tabular}{c}

\renewcommand{\arraystretch}{0.9}
\begin{tabular}{ccccccc} 
\multicolumn{7}{c}{\fontsize{12}{12}\selectfont(c) Evaluate on the unseen dataset~\cite{li2023gait}.} \\
\toprule[2pt]
\multirow{2}{*}{DroneGait~\cite{li2023gait}} & \multicolumn{3}{c}{Mid-pitch views} & \multicolumn{3}{c}{High-pitch views}  \\ 
 \addlinespace[-0.5mm] \cmidrule(lr){2-4} \cmidrule(lr){5-7} \addlinespace[-0.5mm]
                         & NM   & BG   & CL                                        & NM   & BG   & CL                                          \\ 
\hline 
GaitSSB-0.03B            & 91.1 & 85.5 & 51.4                                      & 34.8 & 25.9 & 17.6                                        \\ 
GaitSSB-0.13B            & 89.0 & 82.5 & 50.6                                      & 36.0 & 27.1 & 17.8                                        \\ 
FoundationGait-0.03B     & 95.3 & 90.5 & 57.8                                      & 36.8 & 26.9 & 17.7                                        \\ 
FoundationGait-0.13B     & \textbf{96.9} & \textbf{93.2} & \textbf{63.4}                                      & \textbf{50.4} & \textbf{37.5} & \textbf{24.8}                                        \\
\bottomrule[2pt]
\end{tabular}

\\
\\

\renewcommand{\arraystretch}{0.85}
\begin{tabular}{cccccc} 
\multicolumn{6}{c}{\fontsize{12}{12}\selectfont(d) Extending to the unseen modality.} \\
\toprule[2pt]
Modality                       & Method               & Venue    & Rank-1 & Rank-5 & mAP   \\ 
\hline
\multirow{3}{*}{Sil.}         & VPNet~\cite{ma2024learning}                & CVPR'24  & 75.4   & 87.1   & -     \\ 
                               & DeepGaitV2~\cite{fan2023opengait}           & TPAMI'25 & 74.4   & 88.0   & 65.8  \\ 
                               & GaitMoE~\cite{huang2024occluded}              & ECCV'24  & 73.7   & -      & 66.2  \\ 
\cline{2-6}
                               % & FG-0.03B & ours     & 77.7   & 89.6   & 71.1  \\ 
                               % & FG-0.13B & ours     & 79.3   & 89.5   & 74.0  \\ 
\hline
\multirow{2}{*}{Parsing}       & FoundationGait-0.03B & ours     & 82.0   & 91.6   & 76.1  \\ 
                               & FoundationGait-0.13B & ours     & 83.8   & 92.7   & 79.3  \\ 
\hline
\multirow{4}{*}{\begin{tabular}[c]{@{}c@{}}Sil. \\+Parsing\end{tabular}} & XGait~\cite{zheng2024takes}                & MM'24        & 80.5   & 91.9   & 73.3  \\ 
                               & MultiGait++~\cite{jin2025exploring}          & AAAI'25  & 85.4   & \textbf{94.9}   & 80.5  \\ 
\cline{2-6}
                               & FoundationGait-0.03B & ours     & 84.2   & 92.8   & 78.8  \\ 
                               & FoundationGait-0.13B & ours     & \textbf{86.5}   & 93.8   & \textbf{82.6}  \\
\bottomrule[2pt]
\end{tabular}

\end{tabular}

\end{tabular}

}
\label{tab: part-aware ablation}
\label{tab: part-awared on Supervised Method}
\label{tab: ViT}
\label{tab: cross modality}
\end{table}

\noindent \textbf{Diversity vs. Scale.}
To separate the effects of diversity and scale, we subsample WebGait-2M to 1M ($1/2.5$ per dataset), matching GaitLU-1M.
At equal scale (Tab.~\ref{tab:diversity} (a)), both methods favor WebGait-1M: GaitSSB-0.13B by 6.6\% and FoundationGait-0.13B by 19.5\% on average.
This confirms that diversity, not scale alone, is critical for large-scale gait learning.

\noindent \textbf{Head Initialization.}
By default we initialize the head via linear probing before fine-tuning, akin to two-stage tuning in VLMs.
As shown in Tab.~\ref{tab:head-init} (b), this slightly outperforms fine-tuning from scratch (+0.3\% F1), though both remain competitive, indicating robustness to head initialization.

\begin{table}[t]
\centering
\caption{\textbf{Ablation Study.} 
(a) Effect of pretraining data diversity at matched 1M scale.
(b) Effect of head initialization before fine-tuning.
}
\resizebox{0.8\columnwidth}{!}{

\begin{tabular}{c}

\begin{tabular}{ccccccccc}
\multicolumn{9}{c}{\fontsize{12}{12}\selectfont(a) Effect of pretraining data diversity at matched 1M scale.} \\
\toprule[2pt]
\multirow{2}{*}{\begin{tabular}[c]{@{}c@{}}Source\\Dataset\end{tabular}} & \multirow{2}{*}{Method} & OU-MVLP    & \multicolumn{2}{c}{Gait3D}   & \multicolumn{4}{c}{CCPG} \\
\cmidrule(lr){3-3} \cmidrule(lr){4-5} \cmidrule(lr){6-9}
                                                                          &                         & Rank-1  & Rank-1 & Rank-5 & CL   & UP   & DN   & BG   \\
\hline
\multirow{2}{*}{\begin{tabular}[c]{@{}c@{}}GaitLU-1M\end{tabular}}
  & GaitSSB-0.13B                    & 26.3          & 25.1 & 39.9 & 11.1 & 20.0 & 31.0          & 49.4 \\
  & FoundationGait-0.13B             & 35.5          & 26.6 & 40.1 & 12.3 & 23.0 & 32.1 & 53.4 \\
\hline
\multirow{2}{*}{\begin{tabular}[c]{@{}c@{}}WebGait-1M\end{tabular}}
  & GaitSSB-0.13B                    & 34.5          & 30.2          & 48.4          & 14.7 & 22.5 & 41.1 & 57.9 \\
  & FoundationGait-0.13B             & \textbf{58.6} & \textbf{42.5} & \textbf{60.1} & \textbf{24.3} & \textbf{35.3} & \textbf{60.8} & \textbf{77.9} \\
\bottomrule[2pt]
\end{tabular}

\\

\begin{tabular}{ccccccc} 
\\
\multicolumn{7}{c}{\fontsize{10}{10}\selectfont(b) Effect of head initialization before fine-tuning.} \\
\toprule[2pt]
Mode & Head Initialization & Method              & Acc.                    & Precision              & Recall        & F1                      \\ 
\hline
\multirow{2}{*}{\begin{tabular}[c]{@{}c@{}}Fine-\\tuning\end{tabular}} & Scratch & FoundationGait-0.13B & 94.4 & \textbf{83.0} & 94.5 & 84.8 \\ 
\cmidrule(lr){2-7}
& Linear-probing & FoundationGait-0.13B & \textbf{96.0} & 80.1 & \textbf{97.5} & \textbf{85.1} \\ 
\bottomrule[2pt]
\end{tabular}

\end{tabular}
}
\label{tab:diversity}
\label{tab:head-init}
\end{table}

\section{Challenges and Limitations}
\label{sec:conclusion}
We introduce FoundationGait, a scalable self-supervised gait foundation model that addresses key challenges in size scalability and cross-task generalization. 
FoundationGait shows that a unified model can effectively tackle both recognition and healthcare tasks, paving the way for more practical gait applications.

In the end, three issues warrant further investigation:

\noindent \textbf{ViT-based Gait Foundation Model}. 
This issue is crucial for bridging gait-specific foundation models with those from other domains, particularly LLMs~\cite{vaswani2017attention}, yet it remains largely unexplored within the gait community. 

\noindent \textbf{FoundationGait-1B}. 
Due to computational constraints, we only scaled the model up by 20$\times$ to 0.13B and verified that its performance still follows the scaling trends in Fig.~\ref{fig:unscalable}.
Can we scale it beyond 1B? 
This is an ongoing goal.

\noindent \textbf{Toward RGB.}
While silhouette-based gait methods remain dominant in top venues (62.8\% \textit{vs.} 5.1\% for RGB~\cite{fan2025opengait}), the performance and generalization advantages of RGB-based approaches are becoming increasingly evident. 
We hope this work serves as a principled foundation and source of inspiration for the design of RGB-based gait foundation models in the future.

\bibliographystyle{splncs04}
\bibliography{main}
\end{document}